\documentclass[10pt,twocolumn,letterpaper]{article}

\usepackage[pagenumbers]{cvpr} 

\usepackage[accsupp]{axessibility}  

\usepackage{graphicx}
\usepackage{amsmath}
\usepackage{amssymb}
\usepackage{booktabs}

\usepackage[utf8]{inputenc}
\usepackage[T1]{fontenc}
\usepackage{url}
\usepackage{makecell}
\usepackage{amsfonts}
\usepackage{nicefrac}
\usepackage{microtype}
\usepackage{xcolor}
\usepackage{lipsum}
\usepackage{multirow}
\usepackage{tabularray}

\newcommand{\gray}[1]{\textcolor{gray}{{#1}}}

\definecolor{cvprblue}{rgb}{0.21,0.49,0.74}
\usepackage[pagebackref,breaklinks,colorlinks,allcolors=cvprblue]{hyperref}

\def\paperID{EDGE-31} 
\def\confName{CVPR}
\def\confYear{2025}

\author{Chaitanya Patel \quad\quad
Juan Carlos Niebles \quad\quad
Ehsan Adeli \\ 
Stanford University \\
{\small\url{https://chaitanya100100.github.io/AdaVid/}}
}

\begin{document}

\title{AdaVid: Adaptive Video-Language Pretraining}
\maketitle
\begin{abstract}
Contrastive video-language pretraining has demonstrated great success in learning rich and robust video representations. However, deploying such video encoders on compute-constrained edge devices remains challenging due to their high computational demands. Additionally, existing models are typically trained to process only short video clips, often limited to 4 to 64 frames. In this paper, we introduce AdaVid, a flexible architectural framework designed to learn \textit{efficient} video encoders that can dynamically adapt their computational footprint based on available resources. At the heart of AdaVid is an adaptive transformer block, inspired by Matryoshka Representation Learning, which allows the model to adjust its hidden embedding dimension at inference time. We show that AdaVid-EgoVLP, trained on video-narration pairs from the large-scale Ego4D dataset, matches the performance of the standard EgoVLP on short video-language benchmarks using only half the compute, and even outperforms EgoVLP when given equal computational resources. We further explore the trade-off between frame count and compute on the challenging Diving48 classification benchmark, showing that AdaVid enables the use of more frames without exceeding computational limits. To handle longer videos, we also propose a lightweight hierarchical network that aggregates short clip features, achieving a strong balance between compute efficiency and accuracy across several long video benchmarks.

\end{abstract}
\section{Introduction}

\begin{figure}[t]
\centering
\includegraphics[width=0.40\textwidth]{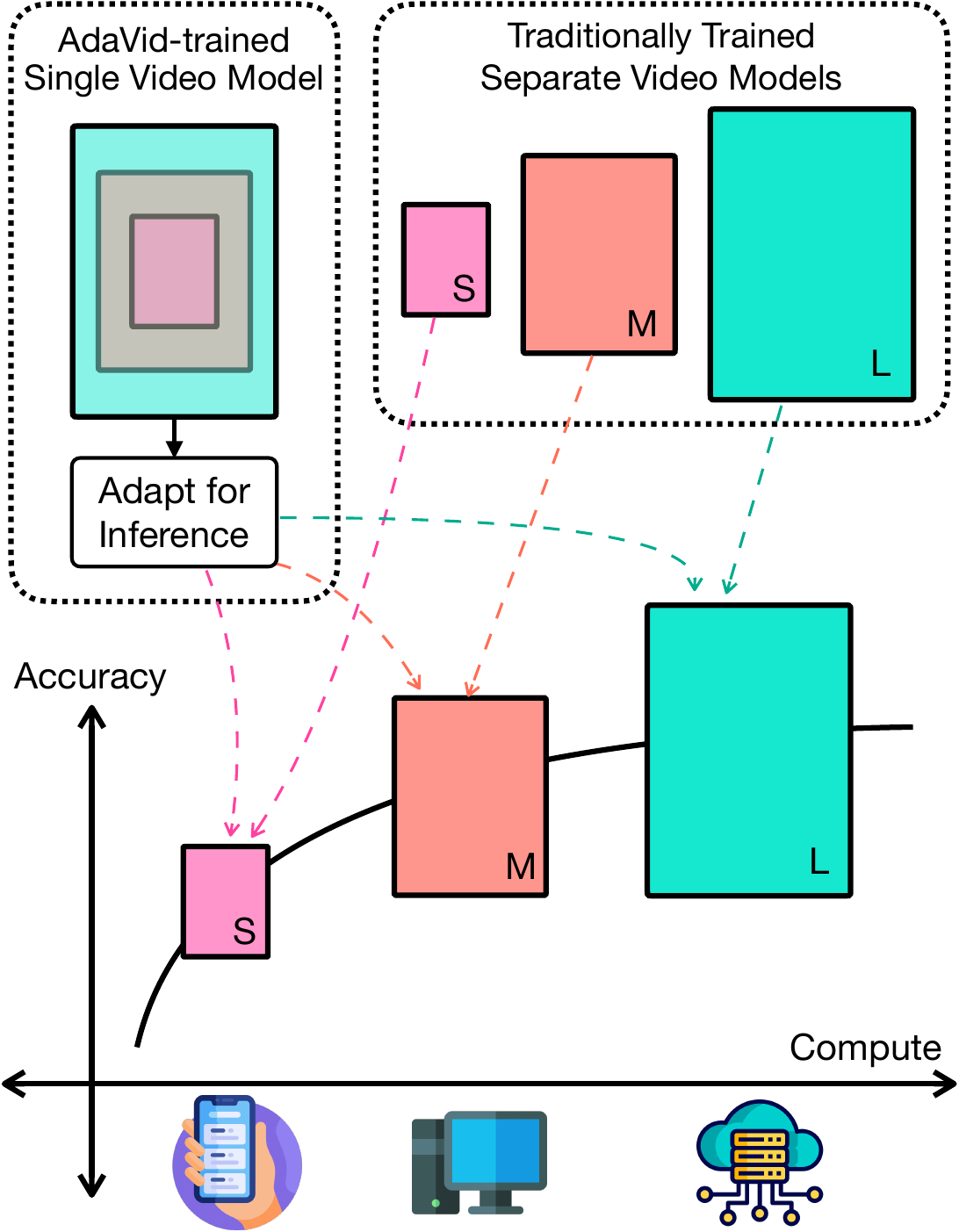}
\caption{A single AdaVid-trained video model facilitates inference with controllable computational footprint without any postprocessing. It allows one model to adjust its computational demands dynamically according to the requirements, thereby eliminating the need to train multiple distinct models.}
\label{fig:teaser}
\end{figure}

Image-language pretraining~\cite{CLIP} has shown remarkable success in learning rich image representations that are robust and transferable to multiple downstream tasks.
Inspired by this success, video-language models~\cite{SpaceTimeAttn,Frozen,EgoVLP,VideoCLIP} have emerged as a promising direction to learn rich video representations that are transferable to downstream tasks such as video-text retrieval, video question answering, action recognition etc.
Typically, both video and text encoders are transformer-based architectures~\cite{DistilBERT,Frozen,VideoCLIP} where compute and memory requirement increases quadratically with the input number of tokens.
Video encoders are especially compute-inefficient because even a small number of frames results in a very high number of tokens.
Consequently, these models are trained by sampling a small number of video frames (typically 4 to 16).
This computational and data inefficiency becomes prohibitive when attempting to train long-form video encoders, especially under contrastive learning frameworks that require larger batch sizes to learn better features.
This limitation also restricts their deployment on edge devices with constrained computational resources.

Several prior works have focused on developing efficient transformer architecture, particularly aiming to address the quadratic complexity of self-attention~\cite{beltagy2020longformer,guan2022transkimmer,jaegle2021perceiver}.
Many works leverage the redundancy and structure of video input through space-time attention~\cite{SpaceTimeAttn}, hierarchical modeling~\cite{HierVL}, or memory-based architectures~\cite{mcvit}.
In this work, we draw inspiration from the fact that the computational complexity of transformer is also quadratic with respect to the token dimension and propose AdaVid, an architectural framework to encode long videos in an efficient and adaptive manner.
The key component of AdaVid is an adaptive transformer block, inspired by Matryoshka Representation Learning (MRL)~\cite{MRL}, that can process input tokens of varying dimensions by sampling appropriate parameters.
This design offers the flexibility to dynamically adjust the embedding size of each transformer layer during inference.
As shown in ~\Cref{fig:teaser}, one AdaVid-trained video encoder encompasses multiple models of different capacities, enabling the encoding of both long and short videos while accommodating a flexible compute footprint.

To show the effectiveness of AdaVid, we train an adaptive version of EgoVLP~\cite{EgoVLP} within our proposed framework (referred to as AdaVid-EgoVLP) on the large-scale Ego4D~\cite{Ego4D} video-language dataset. 
We show that AdaVid-EgoVLP performs equal or better than vanilla EgoVLP and other baselines on several benchmarks while using the full embedding dimension.
Additionally, we also show that AdaVid-EgoVLP retains comparable performance on those benchmarks while operating with low embedding dimension (and hence low compute resources).
We carry out compute vs. accuracy analysis based on varying dimension sizes of different layers, and provide key insights into these design choices.

We also conduct a compute vs. frame count analysis on the Diving48~\cite{li2018diving48} long video classification benchmark and show that AdaVid enables the model to process more frames within limited compute while maintaining strong classification accuracy.
Additionally, we train AdaVid-Agg -- a lightweight aggregator network that can distill the sequence of AdaVid-EgoVLP embeddings extracted from consecutive video clips, into a single feature vector for the entire video.
Such hierarchical design allows us to train long video-language models using relatively smaller datasets with long video annotations.
We show that AdaVid-Agg retains surprisingly high accuracy on several long video benchmarks, even while operating with a fraction of computational resources chosen adaptively at test time.
Such ability to adaptively change the embedding dimension (and consequently the FLOPs) is highly desirable for video understanding in compute-constrained edge devices and wearable devices where the compute load may also vary from time to time.
In short, our contributions are as follows:
\begin{itemize}
\item We propose an adaptive transformer layer capable of processing input tokens with varying dimensions. This design enables a video encoder, composed of these adaptive layers, to perform inference while accommodating different computational requirements.
\item We introduce AdaVid-EgoVLP -- an adaptive variant of EgoVLP and AdaVid-Agg for short and long video understanding, respectively. We demonstrate improved compute vs. accuracy trade-offs across multiple benchmarks and also show favorable frame-count vs. compute trade-off on the Diving48 long video classification benchmark.
\item We investigate different training and evaluation configurations for choosing dimensions of transformer layers. We show that gradually decreasing embedding dimension sizes across layers yields better performance compared to the alternative approach.
\end{itemize}

\section{Related Work}

\textbf{Efficient Deep Learning Models}: 
Several prior works have focused on creating compute-efficient deep learning architectures. Knowledge distillation~\cite{hinton2015distilling,DistilBERT,deit} requires a two stage process to distill the performance of the bigger teacher model into a smaller student model.
Methods like pruning~\cite{lagunas2021block,liu2018rethinkingpruning} and quantization~\cite{shen2020q} have been proposed to improve the inference efficiency of deep learning models, which may potentially lead to a decrease in performance~\cite{choi2016quantlimit}.
Recently, transformers~\cite{vaswani2017attention} have become the predominant architecture for many pretraining tasks (including language~\cite{vaswani2017attention, BERT}, image-language~\cite{CLIP}, and video-language~\cite{VideoCLIP,Frozen,EgoVLP}) due to their ability to leverage larger datasets and learn richer embeddings from tokenized inputs.
However, the computational complexity of the attention mechanism scales quadratically with the number of input token, potentially rendering current efficiency techniques insufficient.
This presents a significant obstacle when deploying these models in compute-limited environments.
Several prior works have explored methods to improve transformer efficiency, including sub-quadratic attention~\cite{beltagy2020longformer}, token dropping~\cite{guan2022transkimmer}, and token resampling~\cite{jaegle2021perceiver}.
These approaches, however, often sacrifice accuracy for \textit{permanent} compute efficiency.
In contrast, AdaVid maintains compute efficiency when necessary while preserving the accuracy of a standard transformer when operating at full capacity.

\textbf{Adaptive Models}:
Adaptive training to obtain multiple models from a single trained model have been explored in the context of CNNs~\cite{yu2018slimmable,cai2019once,grimaldi2022dynamic} and transformers~\cite{chavan2022vitslim,hou2020dynabert}.
Matryoshka Representation Learning~\cite{MRL} proposed to use adaptive dimensions at the final feature vector, while FlexViT~\cite{beyer2023flexivit} introduced vision transformer with flexible input space. 
Although these approaches simplify training, they require the network backbone to operate at full capacity, offering no computational benefits during inference.
MatFormer~\cite{MatFormer} proposed to use adaptive computation only for FFN layers and projected tokens back to full dimension size for self-attention layers.
Since quadratic self-attention is carried out with full token dimension, the computational benefit of this design remains limited, particularly in video understanding tasks where the number of tokens is high.
For language modeling, SHARCS~\cite{SHARCS} proposed to use a separate router network to predict the difficulty of a sample, requiring a separate heuristic-based computation of sample hardness during every training epoch.
More recently, ~\cite{zhang2024poa} proposed to use a third elastic student network in DINO~\cite{caron2021dino} image pretraining and used multiple cross-view distillation losses between student and teacher models.
In contrast, we simplify the design by incorporating an adaptive embedding dimension at every transformer layer, without the need for distillation and heuristic calculations of sample difficulty.

\textbf{Video-Language Models}:
Video-language pretraining~\cite{VideoCLIP,Frozen,EgoVLP,miech2019howto100m,miech2020end,luo2020univl,lin2022egocentricpretraining} has become a key method for developing rich video embeddings for various downstream tasks.
Since the video encoders have high memory and compute footprint, these models are typically trained on short videos by sampling few (4 to 64) frames.
For example, EgoVLP~\cite{EgoVLP} trains on video-narration pairs of Ego4D~\cite{Ego4D} dataset by sampling 4 frames per video sample.
This limits the applicability of such methods on long-form video understanding and compute-constrained environments.
Many prior works have proposed video specific solutions to reduce the compute requirements.
Slow-fast networks~\cite{feichtenhofer2019slowfast} uses two separate networks to process video at different frame-rate whereas sampling-based methods~\cite{liu2018rethinkingpruning,zhi2021mgsampler,wang2024videoagent} samples few informative frames.
To avoid quadratic global attention, some works propose efficient attention mechanism~\cite{SpaceTimeAttn,wang2022deformable,islam2022statespacevideo}.
Memory-based architectures~\cite{wu2022memvit,mcvit} processes videos in a streaming manner with a memory mechanism to store key information of the past.
HierVL~\cite{HierVL} employs a hierarchical model where a video is broken into small video clips and encoded with a standard video encoder~\cite{EgoVLP}.
A separate small aggregator network is employed to aggregate segment features into a long video feature.
Although this formulation allows HierVL to train with long videos (up to 64 frames), it compromises its accuracy on short videos in favor of long videos.

Our AdaVid framework for efficient video understanding is orthogonal to these works.
AdaVid focuses on training a model with adaptive compute for compute-contrained edge devices and wearable devices.
AdaVid can leverage the redundancy in video inputs and learn rich video embeddings with a fraction of compute compared to baselines.
We show applicability of AdaVid on a video encoder with space-time attention~\cite{EgoVLP} as well as hierarchical modeling~\cite{HierVL}.
\section{Method}

\begin{figure*}[t]
\centering
\includegraphics[width=0.95\textwidth]{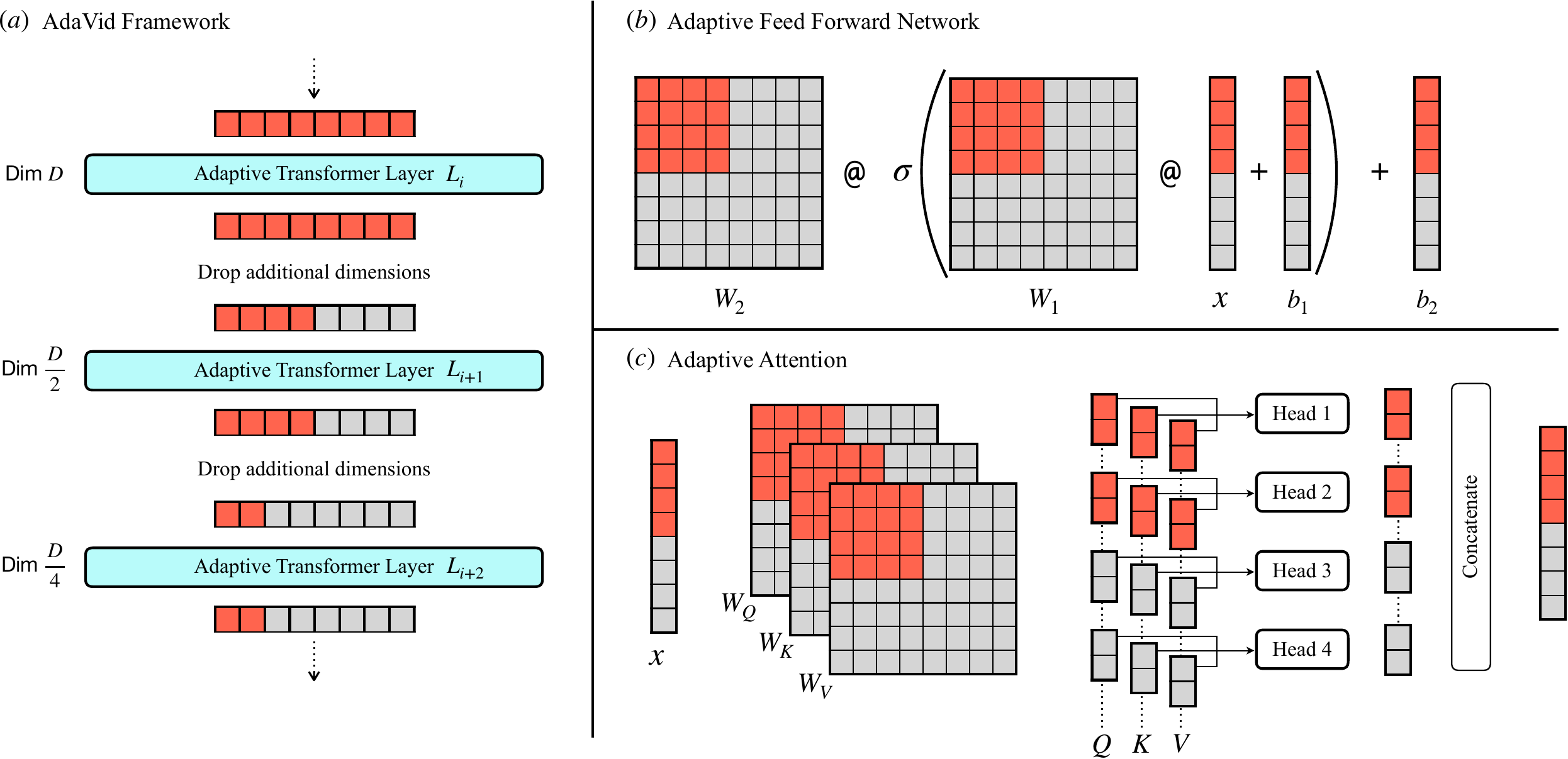}
\caption{\textbf{AdaVid Framework} is designed to train video encoders that facilitate adaptive compute-efficient inference.
(a) Key component of AdaVid is the Adaptive Transformer Layer, which is designed to handle input tokens of varying dimension sizes up to $D$. During each training iteration, each layer processes the input tokens with a randomly selected dimension size, enforcing a coarse-to-fine structure in the model's weights and activations. This allows an AdaVid-trained model to perform inference with a controllable compute footprint.
(b) The feedforward layer $W_2\;\sigma(W_1x+b_1)+b_2$ of the transformer can be modified to accommodate input tokens of size $D/2$ by appropriately slicing the weight and bias parameters. This approach is also applicable to the affine transformation of layer normalization.
(c) In multi-head attention, input tokens of size $D/2$ are processed using half the number of heads, rather than reducing the dimension of each head.}
\label{fig:method}
\end{figure*}

\subsection{Preliminary}
\label{subsec:prelim}
Video input is denoted as a tensor of size $T\times3\times H\times W$ where $T$ is the input number of frames.
Each frame is divided into patches of size $P\times P$, giving $N=HW/P^2$ patches per frame.
A common practice is to use $H=W=224$ and $P=16$ which gives $N=196$ patches per frame.
Each patch of each frame is projected into a $D$-dimensional space using a linear projector, giving a total $TN$ number of tokens of dimension $D$.
At every transformer layer, these tokens go through self-attention and FFN layers.
The complexity of every FFN layer is usually $16ND^2$.
See the Appendix for a detailed description of FLOPs computation.

\textbf{Dense Attention}: If each transformer layer were to use vanilla full attention, its compute complexity would be $24TND^2+4T^2N^2D$ (See Appendix).
Typically, a \texttt{[cls]} token is appended to represent the video embedding, but since $TN >> 1$, we ignore it here for brevity.
Note that this formulation is quadratic in $N$, $T$, and $D$.

\textbf{Space-Time Attention}: To avoid calculating dense attention, TimeSformer~\cite{SpaceTimeAttn} introduced divided space-time attention where each token first attends to other tokens of the same frame and then attends to the same-positioned tokens of other frames.
The compute complexity of each space-time transformer layer is $32TND^2+4TND(N+T)$ (See Appendix).
Since its complexity scales linearly with the number of frames $T$, it has been a preferred choice for many subsequent video encoders like Frozen~\cite{Frozen}, EgoVLP~\cite{EgoVLP}, etc.
However, because of the high constant of the first term, its relative computational benefits are realized only at a high ($>32$) number of frames which is usually not the case for most video understanding tasks.

\textbf{Hierarchical Modeling}:
To train with longer videos, some works~\cite{HierVL} propose to use hierarchical modeling where the video is divided into $S$ segments of $T/S$ frames each.
Each segment is encoded independently by a standard video encoder (like EgoVLP) to give $S$ segment features.
This sequence of $S$ segment features is then aggregated by another network to output a single feature embedding for the long video.
Such hierarchical modeling makes sense from the perspective of data-efficiency because video modality has a lot of redundancy in pixel values and distant frames relate to each other only through high-level semantic concepts (Dense attention or even space-time attention would be excessive in such cases).
However, it does not provide notable computational benefit over space-time attention because the effective complexity of each layer of the feature extractor is $32TND^2+4TND(N+T/S)$ (See Appendix).
The complexity of the hierarchical aggregator is relatively negligible and can be ignored.

\subsection{Adaptive Transformer Layer}

Note that the complexity of every transformer component is also quadratic with $D$, often with a high constant. Building on this insight, we propose to train video encoders that can adaptively use smaller embedding dimension at every transformer layer during inference. 

Consider a transformer layer that processes a sequence of $K$ tokens $(x_1, \cdots, x_K)$, producing $K$ output tokens $(y_1, \cdots, y_K)$, where each $x_i, y_i \in \mathbb{R}^D$.
As shown in ~\Cref{fig:method}, we adapt its components to handle tokens of smaller dimension $d\leq D$ so that a trained adaptive transformer layer can process a sequence of smaller tokens $(\hat{x}_1, \cdots, \hat{x}_K)$, where each $\hat{x}_i\in \mathbb{R}^d$ consists of the first $d$ values of  $x_i$. It outputs $(\hat{y}_1, \cdots, \hat{y}_K)$, where $\hat{y}_i\in \mathbb{R}^d$ retains as much semantic information as $y_{i[1{:}d]} \in \mathbb{R}^d$ where $y_{i[1{:}d]}$ is a vector of first $d$ values of $y_i$.

This is achieved by making every basic component of the transformer adaptive. For FFN, every linear projection of form $y=W\cdot x+b$, can be adjusted to $y_{[1{:}d]}=W_{[1{:}d{,}1{:}d]}\cdot x_{[1{:}d]} + b_{[1{:}d]}$, using the upper-left $d\times d$ submatrix of weight matrix $W$ and first $d$ values of bias vector $b$.
Similarly, layer normalization $\text{LN}(x;\gamma,\beta)$ can be adjusted to $\text{LN}(x_{[1{:}d]};\gamma_{[1{:}d]},\beta_{[1{:}d]})$. For multi-head attention, instead of reducing the dimension for each head, we reduce the number of heads~\cite{SHARCS}. In particular, if the vanilla transformer layer has $D/H$ heads each with dimension $H$, the adaptive transformer layer uses $d/H$ heads. To incorporate this, we only use $d$ which is a multiple of $H$ in our experiments.


\subsection{AdaVid Video-Language Pretraining}

We present AdaVid as a general architectural framework where every transformer layer of any standard video encoder can be replaced with our adaptive transformer layer.
In this paper, we show the effectiveness of AdaVid on contrastive video-language representation learning for short and long videos.
Specifically, we introduce AdaVid-EgoVLP and AdaVid-Agg for encoding short and long videos, respectively.
These models leverage our adaptive transformer block to encode videos in a compute-adaptive manner.

\textbf{AdaVid-EgoVLP}: For short videos, we follow the exact setup of EgoVLP~\cite{EgoVLP} and train its adaptive counterpart AdaVid-EgoVLP.
Its video encoder uses $T=4$ frames, image size $H=W=224$, and patch size $P=16$ to tokenize the video clip input.
These tokens are processed by 12 adaptive transformer layers with a maximum dimension size of $D=768$, each consisting of an adaptive space-time attention~\cite{SpaceTimeAttn} module followed by an adaptive feedforward network.
We use DistilBERT~\cite{DistilBERT} as our text encoder and finetune it during our experiments.
Following EgoVLP, we also use EgoNCE~\cite{EgoVLP} loss to train AdaVid-EgoVLP.
We compare AdaVid-EgoVLP with vanilla EgoVLP and other strong baselines on short video benchmarks and carry out compute vs. accuracy analysis.


\textbf{AdaVid-Agg}: For long videos, we follow hierarchical late fusion modeling and train a lightweight AdaVid-Agg model to aggregate a sequence of consecutive video clip features extracted from AdaVid-EgoVLP.
In particular, we sample $T=64$ frames from the input long video (unless mentioned otherwise) and encode $S=16$ segments (each containing 4 consecutive frames) independently using AdaVid-EgoVLP.
AdaVid-Agg aggregates $S$ segment features into a single feature for the long video.
AdaVid-Agg is implemented as a transformer~\cite{vaswani2017attention} consisting of 12 transformer layers.
Since it operates over a short sequence of segment features instead of patch tokens, its compute footprint is negligible compared to the compute required for AdaVid-EgoVLP feature extraction.
Note that our setup is simpler than HierVL~\cite{HierVL} which requires large-scale multi-node training to jointly learn aggregator and EgoVLP feature extractor and uses specific losses based on the hierarchical annotations.
In contrast, we do not require hierarchical annotations and train aggregator independently from the feature extractor using standard InfoNCE~\cite{CLIP} loss, and show comparable performance to HierVL.
We compare AdaVid-Agg with strong baselines on multiple long video benchmarks.


\begin{table}[t]
\footnotesize
\centering
\caption{AdaVid-EgoVLP evaluation configurations. We evaluate the trained AdaVid-EgoVLP model with different configurations for embedding dimensions. [$768 \times 12$] indicates that all 12 layers use 768-d tokens. [$768 \times 4$,  $576 \times 4$, $384 \times 4$] means that first four layers use 768-d, followed by four layers of 576-d, followed by final four layers of 384-d. The FLOPs are computed using the complexity equations provided in \Cref{subsec:prelim} with $T=4$.}
\begin{tabular}{clc}
\toprule
Config. & Layer-wise hidden dimension & \makecell{FLOPs\\($\times 10^{10}$)} \\
\midrule
 d-768 & [$768 \times 12$] & 18.3 \\
 d-576 &  [$576 \times 12$] & 10.4 \\
 d-384 &  [$384 \times 12$] & 4.7 \\
 d-192 &  [$192 \times 12$] & 1.2 \\
\midrule
 d-dec & [$768 \times 3$,  $576 \times 3$, $384 \times 3$, $192 \times 3$] & 8.7 \\
 d-dec-high &  [$768 \times 4$,  $576 \times 4$, $384 \times 4$] & 11.2 \\
 d-dec-low &  [$576 \times 4$, $384 \times 4$, $192 \times 4$] & 5.5 \\
\midrule
 \gray{d-inc} &  \gray{[$192 \times 3$, $384 \times 3$,  $576 \times 3$, $768 \times 3$]} & \gray{8.7} \\
 \gray{d-inc-high} &  \gray{[$384 \times 4$, $576 \times 4$, $768 \times 4$]} & \gray{11.2} \\
 \gray{d-inc-low} &  \gray{[$192 \times 4$, $384 \times 4$, $576 \times 4$]} & \gray{5.5} \\
\bottomrule
\end{tabular}
\label{tab:eval_configs}
\end{table}

\textbf{AdaVid Training}: During each iteration of training, an embedding dimension can be chosen for every adaptive transformer layer randomly or based on some strategy.
For our experiments, we fix the set of allowed embedding dimensions to be $\{D, 3D/4, D/2, D/4\}$ where $D$ is the full embedding size.
If the chosen embedding size is higher than the previous embedding size, we pad with zeros; if it is lower, then we simply drop additional dimensions.
AdaVid training can also be viewed as a version of dropout which can provide additional regularization benefits.
It forces a coarse-to-fine structure on the manifold of the latents and model weights which can be stripped short during inference based on need.
Note that our focus is not on performance improvement, and MRL~\cite{MRL} also reported no performance gains over vanilla training.
However, we find that models trained with AdaVid often outperform their vanilla counterparts despite identical training setups, likely due to this regularization effect.

AdaVid allows us to train a single model `containing' multiple smaller models of varying capacities and gives us the flexibility of choosing an embedding dimension at test time.
It is also possible to create a large set of smaller models by choosing different granularity at each transformer layer~\cite{MatFormer}, even the ones not observed during training~\cite{MRL}. 
In our experiment, we show that some strategies of choosing embedding dimension fare better than others.

\section{Experiments}

\subsection{Dataset}
We use large-scale Ego4D~\cite{grauman2022ego4d} dataset which contains 9645 untrimmed videos of varying lengths from 5 sec to 7 hrs, totaling 3000 hours of video data.
These videos contain daily human activities captured from an egocentric perspective using smart glasses.
To train AdaVid-EgoVLP, we use a curated set of $\sim$4M narrations~\cite{EgoVLP} each covering 1-2 seconds of video clip.
To train AdaVid-Agg, we additionally use $\sim$100K summaries~\cite{HierVL} each covering 5 minutes of videos.

\subsection{Evaluation Benchmarks}
We evaluate the quality of video-language embeddings on various zero-shot video-language benchmarks covering short as well as long videos.

\textbf{EgoMCQ}~\cite{EgoVLP}: Given a text description and 5 short video clips (1-2 seconds), classify which video aligns with the text description. It contains $\sim$40K samples. The metric is classification accuracy. It is divided into two subparts: (1) EgoMCQ (inter) where five candidate clips for each text query are sourced from the whole dataset. This is an easier setting. (2) EgoMCQ (intra), where the candidate clips for each text query are sourced from the same video as the correct video clip. This is a harder setting.

\textbf{SummaryMCQ}~\cite{HierVL}: Given a text description and 5 medium-lengthed videos (5 minutes), classify which video aligns best with the text description. The metric is classification accuracy. This is similar to EgoMCQ but for longer 5-minute videos.

\textbf{Diving-48}~\cite{li2018diving48} is a curated action recognition benchmark designed to evaluate long-term temporal reasoning in action recognition. Each video is categorized into one of 48 classes based on the type of dive, requiring fine-grained spatio-temporal reasoning over extended temporal contexts. Unlike many other action recognition datasets, achieving high classification accuracy on Diving-48 requires processing a larger number of frames.

\textbf{LongVideoRetrieval}: As discussed in ~\cite{mangalam2024egoschema}, creating a long video-language benchmark to test models' long video understanding is a challenging task because many video benchmarks can be solved adequately by observing only a few frames. To evaluate the model on even longer videos, we curate LongVideoRetrieval benchmark by using long video captions of Ego4D videos provided by ~\cite{videorecap}. The objective is to retrieve the corresponding video, based on a given long textual description, from a database of approximately 1500 long videos, which range in length from a few minutes to two hours, with an average duration of 29 minutes. Processing a small number of video frames is not enough for LongVideoRetrieval because the captions are long and cover activities happening throughout the video. The evaluation metric is recall (R@1, R@5 and R@10). 

\textbf{EgoSchema}~\cite{mangalam2024egoschema}: Given a 3-minute video and a complex question with 5 choices, predict the correct answer. This benchmark was manually curated for long-form video understanding to make sure that the correct answer cannot be derived by observing a short video clip from the entire video. The metric is classification accuracy. Note that this is a VideoQA benchmark with much more complex language than our pretraining dataset.

\subsection{AdaVid-EgoVLP}

\begin{figure}[t]\centering\footnotesize
\centering
\includegraphics[width=0.35\textwidth]{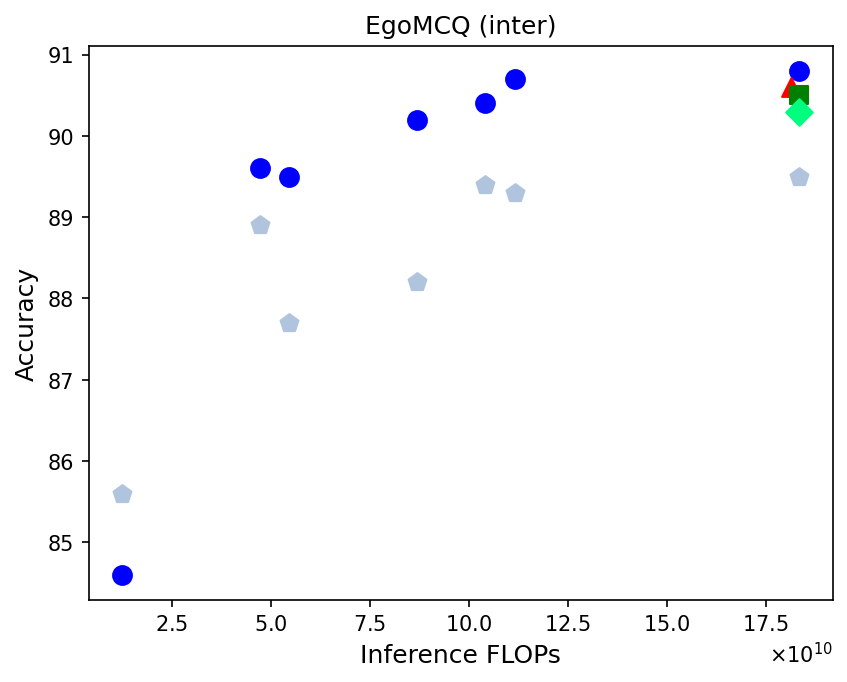}
\includegraphics[width=0.35\textwidth]{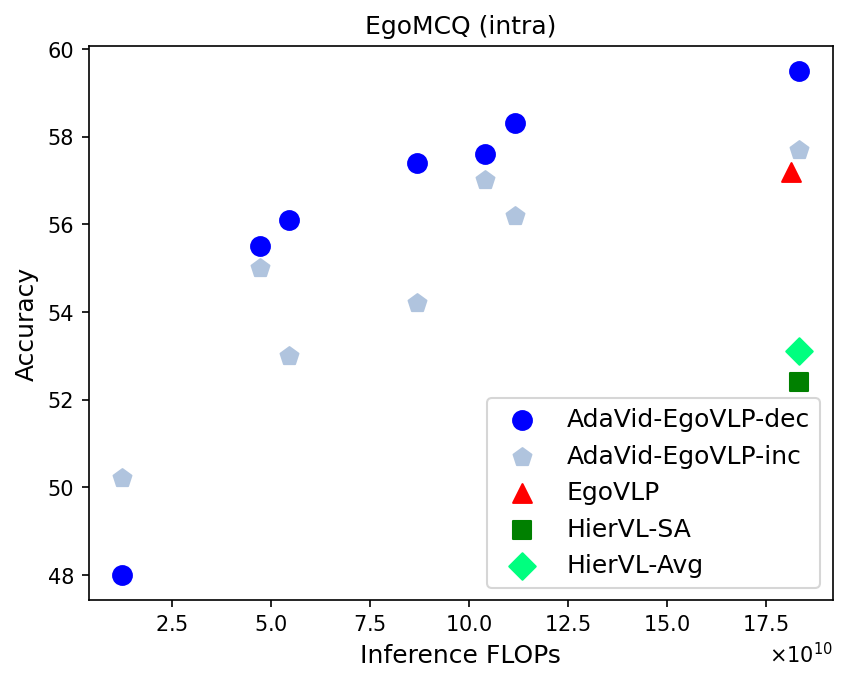}
\caption{\textbf{AdaVid-EgoVLP on two EgoMCQ benchmarks}: AdaVid-EgoVLP-dec, trained with decreasing dimensions for deeper layers, performs better than AdaVid-EgoVLP-inc which was trained with increasing dimensions. AdaVid-EgoVLP-dec performs better than baselines while using maximum compute resources. The same model also retains high accuracy when evaluated with low compute evaluation configurations from ~\Cref{tab:eval_configs}.}
\label{fig:egomcq_plots}
\end{figure}

\begin{table}[t]
\centering \footnotesize
\vspace{2mm}
\caption{Results on EgoMCQ benchmark}
\vspace{-2mm}
\begin{tabular}{lcc}
\toprule
Method & EgoMCQ (inter) & EgoMCQ (intra) \\
\midrule
EgoVLP~\cite{EgoVLP} & 90.6 & 57.2 \\
HierVL-Avg~\cite{HierVL} & 90.3 & 53.1 \\
HierVL-SA~\cite{HierVL} & 90.5 & 52.4 \\
AdaVid-EgoVLP & \textbf{90.8} & \textbf{59.5} \\
\bottomrule
\end{tabular}
\label{tab:egomcq_table}
\end{table}

We train AdaVid-EgoVLP for 10 epochs on 8 NVIDIA L40S GPUs with a total batch size of 160, and other hyperparameters the same as EgoVLP.
To compensate for our lower batch size compared to EgoVLP (160 vs. 512), and to speed up AdaVid training, we initialize our model with EgoVLP weights and finetune it with adaptive embedding dimensions.
We train a single AdaVid-EgoVLP model and evaluate it using different configurations with different compute requirements as mentioned in ~\Cref{tab:eval_configs}.
Apart from standard configurations with the same embedding dimension for all layers (d-768, d-576, d-384, d-192), we also evaluate `d-dec*' configurations where the embedding dimensions decrease for deeper layers and `d-inc*' where the embedding dimensions increase for deeper layers.

\textbf{Choosing Dimensions for AdaVid training}: To find the optimal strategy for varying layer dimensions during training, we trained two versions: (1) AdaVid-EgoVLP-dec where we randomly sample layer dimensions during training ensuring that each layer has an equal or lower dimension than the previous layer, i.e., gradually decreasing dimension sizes. We evaluate this model on standard and decreasing configurations from ~\Cref{tab:eval_configs}. (2) AdaVid-EgoVLP-inc where we sample gradually increasing dimensions during training. We evaluate this model on standard and increasing configurations from ~\Cref{tab:eval_configs}. The results on two subsets of EgoMCQ benchmark are shown in ~\Cref{fig:egomcq_plots}. We can see that AdaVid-EgoVLP-dec performs better than AdaVid-EgoVLP-inc in various evaluation settings despite having the same FLOPs. This result is in line with the results of ~\cite{SHARCS} where the authors used adaptive embedding only at deeper layers. It also aligns with our intuition that deeper layers can afford to strip away low-level details and only store high-level concepts in a smaller subspace. The opposite strategy bottlenecks the information at early layers and hurts model performance in a significant manner. We use AdaVid-EgoVLP-dec to carry out the rest of the analysis, and only use standard and decreasing configurations from ~\Cref{tab:eval_configs} for evaluation.

\textbf{AdaVid-EgoVLP is accurate while being efficient.} ~\Cref{fig:egomcq_plots} also compares AdaVid-EgoVLP with baselines.
EgoVLP is trained using the same architecture and supervision, and with a higher batch size. HierVL additionally uses summary supervision to associate low-level narrations with high-level activities.
Despite that, AdaVid-EgoVLP-dec outperforms both baselines by a noticeable margin when using the full embedding size during evaluation as reported in ~\Cref{tab:egomcq_table}.
We attribute this improvement to the implicit dropout-like regularization provided by AdaVid training.
AdaVid-EgoVLP-dec also performs equal to or better than baselines despite using approximately 0.5x FLOPs from various configurations as shown in ~\Cref{fig:egomcq_plots}.
It maintains good performance even with 0.25x FLOPs, but observes a small drop at the lowest configuration with 0.06x FLOPs, especially for EgoMCQ(intra) which requires more fine-grained video understanding.
Although different configurations with comparable FLOPs show almost similar performance, some tasks may benefit from an optimized configuration.
Overall, AdaVid allows a single model to exhibit varying levels of computational efficiency, enabling it to allocate more compute to challenging tasks while maintaining high efficiency for simpler ones.
Such flexibility in a single-trained model is a unique feature of the AdaVid framework.

\begin{figure}[t]\centering\footnotesize
\centering
\includegraphics[width=0.35\textwidth]{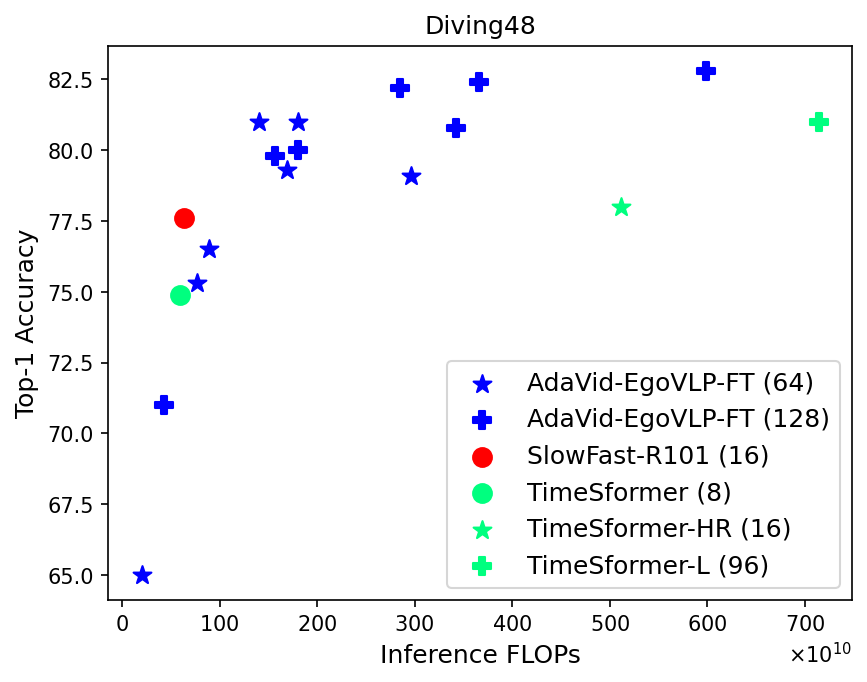}
\caption{\textbf{Results on Diving-48}: We evaluate AdaVid using various evaluation configurations from ~\Cref{tab:eval_configs} with 64 and 128 frames. With adaptive compute, AdaVid can process more frames efficiently, outperforming vanilla-trained baselines.}
\label{fig:diving48_results}
\end{figure}

\begin{table}[t]
\centering \footnotesize
\vspace{2mm}
\caption{Results on Diving-48. The baselines are pretrained on ImageNet-21K, while the AdaVid models are pretrained on Ego4D.}
\vspace{-2mm}
\begin{tabular}{lcccc}
\toprule
Method & \makecell{Num.\\Frames} & \makecell{FLOPs\\$\times 10^{10}$} & \makecell{Top-1\\Accuracy} \\
\midrule
SlowFast-R101~\cite{feichtenhofer2019slowfast} & 16 & 64 & 77.6  \\
TimeSformer~\cite{Frozen} & 8 & 59 & 74.9  \\
TimeSformer-HR~\cite{Frozen} & 16 & 511 & 78.0  \\
TimeSformer-L~\cite{Frozen} & 96 & 714 & 81.0  \\
\midrule
AdaVid-EgoVLP-FT (d-dec) & 64 & 140 & 81.0 \\
AdaVid-EgoVLP-FT (d-dec) & 128 & 285 & \textbf{82.2} \\
\bottomrule
\end{tabular}
\label{tab:d48_table}
\end{table}

\begin{table}[t]
\centering \footnotesize
\vspace{2mm}
\caption{Results on SummaryMCQ and LongVideoRetrieval}
\vspace{-2mm}
\begin{tabular}{lcccc}
\toprule
\multirow{2}{*}{Method} & \multirow{2}{*}{SummaryMCQ} & \multicolumn{3}{c}{LongVideoRetrieval}  \\
\cmidrule{3-5}
 & & R@1 & R@5 & R@10 \\
\midrule
EgoVLP~\cite{EgoVLP} & 89.0 & 7.1 & 21.3 & 31.8 \\
HierVL-Avg~\cite{HierVL} & 95.2 & - & - & - \\
HierVL-SA~\cite{HierVL} & \textbf{95.4} & 16.1 & 48.9 & 62.8 \\
AdaVid-Agg & \textbf{95.4} & \textbf{16.6} & \textbf{50.3} & \textbf{66.5} \\
\bottomrule
\end{tabular}
\label{tab:summarymcq_longvid_table}
\vspace{-2mm}
\end{table}

\begin{figure}[t]
\centering
\includegraphics[width=0.35\textwidth]{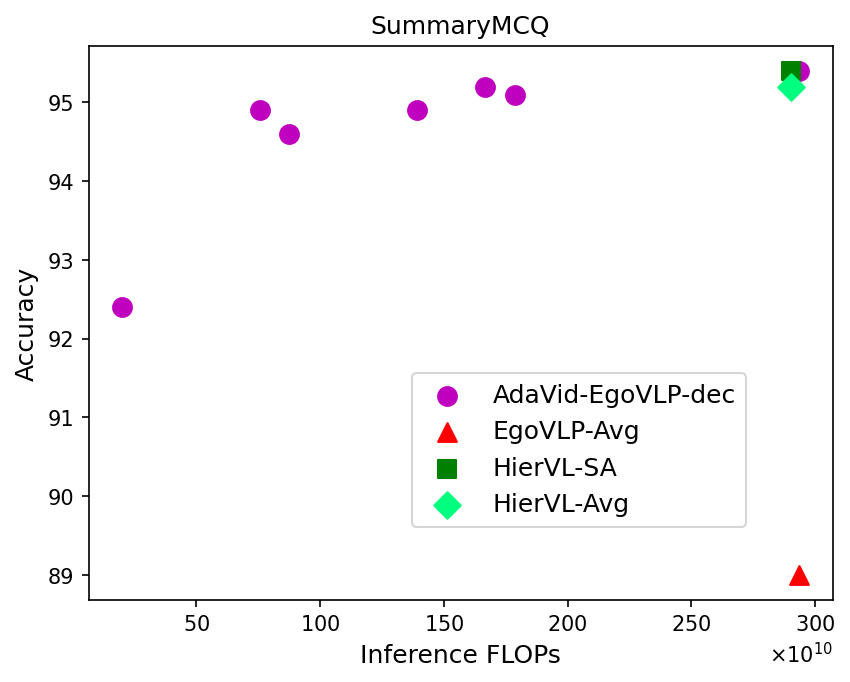}
\caption{\textbf{Results on SummaryMCQ}: AdaVid-Agg achieves comparable performance to HierVL baselines with full embedding dimensions, while also demonstrating robust performance with significantly reduced computational resources as needed.}
\label{fig:summarymcq_results}
\end{figure}

\begin{table}[t]
\centering \footnotesize
\vspace{2mm}
\caption{Results on EgoSchema. Models in gray are pretrained on signficantly larger corpus of video datasets.}
\vspace{-2mm}
\begin{tabular}{lccc}
\toprule
\multirow{2}{*}{Method} & \multirow{2}{*}{\makecell{FLOPs\\$\times 10^{10}$}} & \multicolumn{2}{c}{EgoSchema} \\
\cmidrule{3-4}
& & {Subset} & {Fullset} \\
\midrule
\textcolor{gray}{InternVideo}~\cite{wang2022internvideo} & \textcolor{gray}{2000+} & - & \textcolor{gray}{32.1} \\
\textcolor{gray}{SeViLA}~\cite{yu2024sevila} & \textcolor{gray}{2000+} & \textcolor{gray}{25.7} & \textcolor{gray}{22.7} \\
\textcolor{gray}{LongViViT}~\cite{papalampidi2024longvivit} & \textcolor{gray}{1000+} & \textcolor{gray}{56.8} & \textcolor{gray}{33.3} \\
\textcolor{gray}{MC-ViT-B}~\cite{mcvit}  & \textcolor{gray}{600+} & \textcolor{gray}{61.2} & \textcolor{gray}{42.3} \\ 
HierVL~\cite{HierVL} & 293 & 52.4 & 41.6 \\
\midrule
AdaVid-Agg (d-768) & 293 & 56.2 & 40.9 \\
AdaVid-Agg (d-384)  & 75 & 54.2 & 40.2 \\
AdaVid-Agg (d-192)  & 20 & 52.0 & 38.3 \\
\bottomrule
\end{tabular}
\label{tab:egoschema_table}
\vspace{-2mm}
\end{table}

\textbf{AdaVid can process more frames with limited compute}: 
We finetune AdaVid-EgoVLP on the Diving-48 dataset (referred to as AdaVid-EgoVLP-FT) and evaluate it using different numbers of uniformly sampled frames during inference. ~\Cref{fig:diving48_results} and ~\Cref{tab:d48_table} present the results, compared against TimeSformer~\cite{SpaceTimeAttn}, which shares the same architecture as ours but is trained in a standard (non-adaptive) manner, ensuring a fair comparison. Our results show that AdaVid-EgoVLP-FT can process up to 128 frames using significantly less compute, while outperforming the baseline that relies on much higher computational resources. This shows that when a task demands processing a larger number of frames for temporally fine-grained analysis, AdaVid can effectively scale down the embedding dimension to accommodate more frames within a fixed compute budget.

\subsection{AdaVid-Agg}


\begin{figure*}[t]
\centering
\begin{subfigure}{.33\textwidth}
  \centering
  \includegraphics[width=0.95\linewidth]{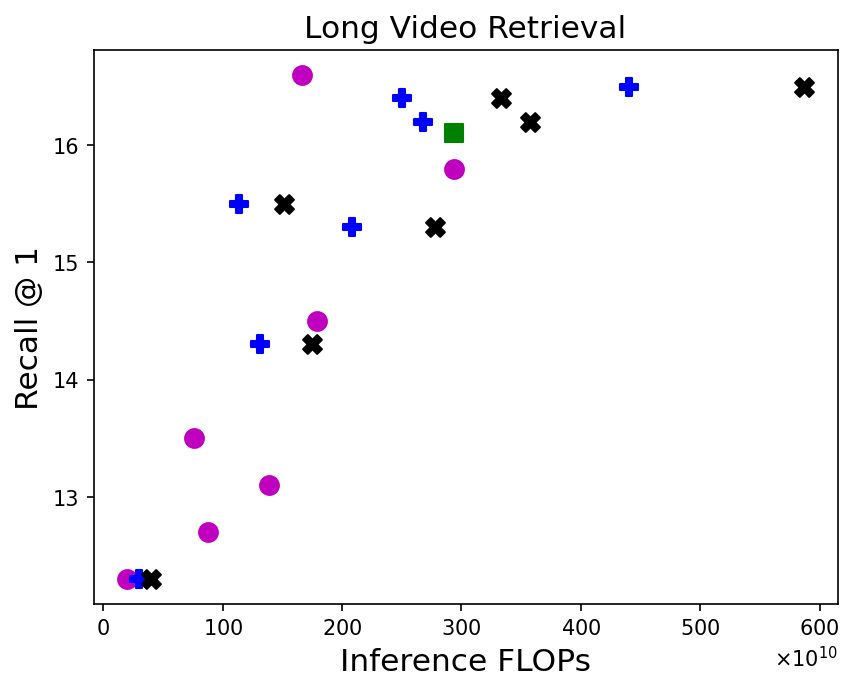}
\end{subfigure}%
\begin{subfigure}{.33\textwidth}
  \centering
  \includegraphics[width=0.95\linewidth]{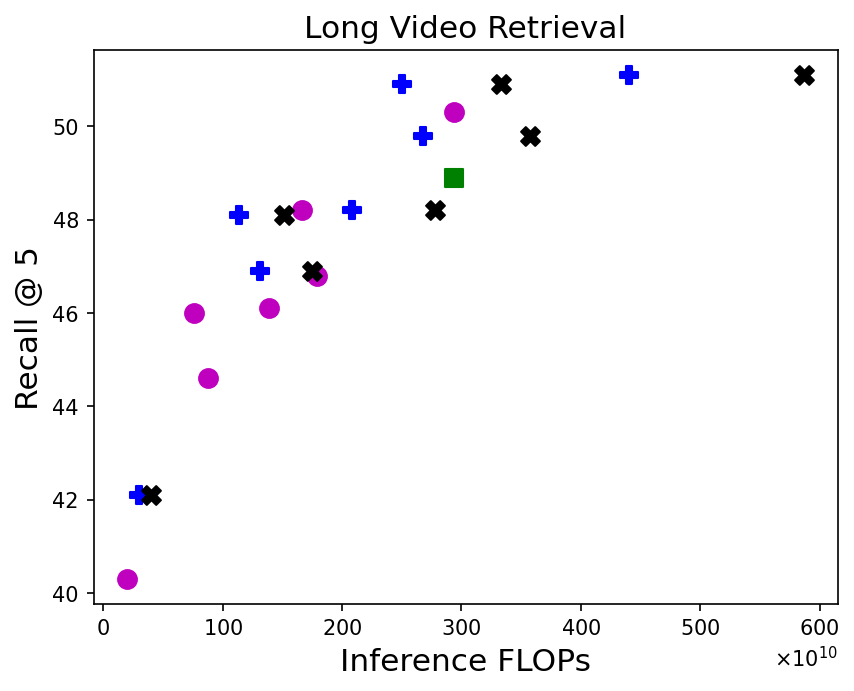}
\end{subfigure}%
\begin{subfigure}{.33\textwidth}
  \centering
  \includegraphics[width=0.95\linewidth]{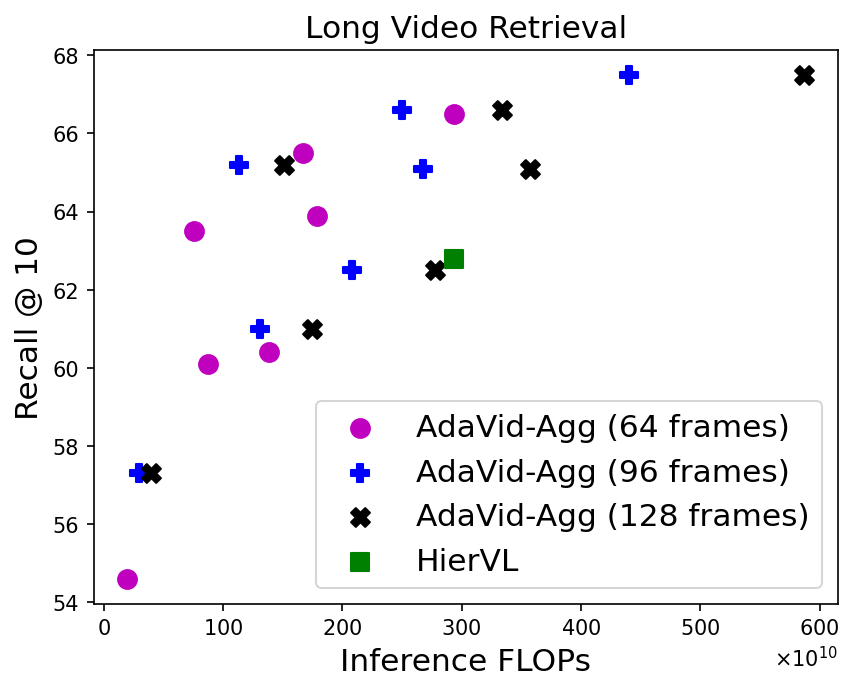}
\end{subfigure}
\caption{\textbf{AdaVid-Agg on LongVideoRetrieval}: We evaluate AdaVid-Agg on text-to-video retrieval from a database of very long videos. We evaluate the dimension configurations mentioned in ~\Cref{tab:eval_configs} using 64, 96, and 128 frames. X-axis shows the compute required to encode a single long video. AdaVid-Agg outperforms HierVL under various inference settings with less compute.}
\label{fig:longvid_results}
\vspace{-2mm}
\end{figure*}

We train AdaVid-Agg for 10 epochs on 8 NVIDIA L40S GPUs on Ego4D short narrations (1-2s) as well as long summary (5 minutes) annotations using a total batch size of 256.
AdaVid-Agg is trained on the extracted features of a pre-trained AdaVid-EgoVLP model where we use different inference configurations provided in ~\Cref{tab:eval_configs} to extract features of varying granularity.
We use a learning rate of $10^{-5}$ and decouple weight decay ~\cite{loshchilov2017decoupled} regularization of 0.1 to account for a relatively smaller video summary dataset.
Similar to before, we train a single AdaVid-Agg model and evaluate it on multiple long video benchmarks.
Since AdaVid-Agg has an extremely small compute footprint relative to the underlying feature extractor AdaVid-EgoVLP, we operate the aggregator at full capacity and carry out compute vs. accuracy analysis by applying different evaluation configurations (~\Cref{tab:eval_configs}) on the feature extractor.

\textbf{Results on SummaryMCQ}: We first evaluate AdaVid-Agg on SummaryMCQ benchmark, which is the long video counterpart of EgoMCQ benchmark, and compare against HierVL baselines in ~\Cref{fig:summarymcq_results}. Similar to AdaVid-EgoVLP, AdaVid-Agg matches the performance of HierVL while using the full embedding dimension as shown in ~\Cref{tab:summarymcq_longvid_table}, and maintains strong performance even with 0.25x compute. EgoVLP-Avg baseline independently encodes $S=16$ short segments of the input video and averages their features. Its significantly worse performance highlights the importance of temporal aggregator network and long video summary supervision in both HierVL and AdaVid-Agg.


\textbf{Results on EgoSchema}:
In ~\Cref{tab:egoschema_table}, we show AdaVid-Agg accuracy on two subsets of EgoSchema benchmark~\cite{mangalam2024egoschema}.
Even though strong HierVL baseline jointly trains feature extractor and aggregator with higher batch size and uses annotation hierarchy to define additional losses, AdaVid-Agg shows similar accuracy as HierVL, and more importantly retains strong performance in low-compute configurations.
For example, when operating AdaVid-EgoVLP with 0.25x compute, it shows very small drop in performance. Even with 0.0625x compute, AdaVid-Agg outperforms many large video models, trained on much bigger corpus of video datasets.
Note that the questions in EgoSchema benchmark has complex language structure in contrast to the simpler narrations and summary annotations of Ego4D on which our method has been trained.
With extensive datasets containing high-quality annotations, AdaVid has the potential to train accurate and dynamically efficient video models suitable for deployment on edge devices.

\textbf{Results on LongVideoRetrieval}:
We evaluate AdaVid-Agg on LongVideoRetrieval benchmark using different evaluation configurations.
In addition to 64 frames, we also evaluate our model using 96 and 128 frames, and show our results in ~\Cref{fig:longvid_results}.
Overall, our single model shows progressively improving retrieval performance when evaluated with compute resources ranging from $0.2$ TFLOPs to $6$ TFLOPs.
AdaVid-Agg outperforms HierVL while using equal or even less compute as shown in ~\Cref{tab:summarymcq_longvid_table}.
It is also possible to do adaptive retrieval~\cite{MRL} where the large set of candidate videos are ranked using an efficient inference configuration to find a smaller set of promising candidates that can be reranked again with more accurate inference with more compute.
We leave this exploration for further research.
\section{Conclusion}

In this paper, we proposed AdaVid framework as a promising direction to learn flexible models that can encompass multitudes of big and small models into a single one.
We showed the effectiveness of AdaVid on video-language pretraining where the video modality has traditionally been compute and data intensive.
Our short video feature extractor, AdaVid-EgoVLP, serves as a flexible replacement for EgoVLP, while AdaVid-Agg is an efficient aggregator of short video segment features.
AdaVid models outperform strong baselines such as HierVL and EgoVLP on both short and long video benchmarks, while also maintaining strong performance under low-compute settings.
We conduct a detailed analysis of accuracy vs. compute and frame-count vs. compute for AdaVid models and baselines, offering valuable insights into leveraging the flexibility of each adaptive layer.
We believe that the AdaVid framework enables the deployment of video-language models on edge devices, supporting efficient long video understanding in a compute-efficient manner.

\noindent\textbf{Acknowledgements}: 
This work was partially supported by Panasonic and the NIH Grant R01AG089169.
We thank Yuta Kyuragi for his helpful feedback on the manuscript.

{
    \small
    \bibliographystyle{ieeenat_fullname}
    \bibliography{main}
}




\clearpage
\maketitlesupplementary
\appendix

\section{FLOPs computation}
\subsection{Multi-Head Self-Attention}
We follow ~\cite{hoffmann2022trainingchinchilla} for calculating the total FLOPs of Multi-head self attention as $8ND^2 + 4N^2D$, where $N$ represents the number of tokens and $D$ denotes the token dimension.

This computation is detailed as follows: The FLOPs required for matrix multiplication of sizes ($M \times N$) and ($N \times P$) is $2MNP$. For each attention head, each projection matrix (query, key, value) involves  $2 \times N \times D \times \frac{D}{H}$ FLOPs. Thus, the total FLOPs for this operation are $6ND^2$.
Each head computes the dot product between the query and key, involving $2 \times N \times \frac{D}{H} \times N$ FLOPs. The total FLOPs for this step amount to $2N^2D$
Each head also computes the weighted sum of values, requiring $2 \times N \times N \times \frac{D}{H}$ FLOPs. Thus, the total FLOPs for this computation are $2N^2D$.
Finally, the output projection layer involves $2 \times N \times D \times D = 2ND^2$ FLOPs. 

\subsection{Feed-forward network}
The token-wise feed-forward (MLP) layer consists of two matrix multiplications: The first transformation converts from dimension $D$ to dimension $F$, while the second converts from dimension $F$ back to dimension $D$. Each matrix multiplication operation involves $2NDF$ FLOPs. Therefore, the total number of FLOPs for the feed-forward network (FFN) is $4NDF$. Typically, the value of F is set to $4D$, resulting in a total of $16ND^2$ FLOPs.

\subsection{Transformer Layers for Video Encoder}
As mentioned in the paper, input video with $T$ frames are patchified where each frame has $N$ tokens, giving total $TN$ number of tokens. These tokens can be processed by multiple transformer layers, each consisting of an attention module and a FFN module. If each layer uses \textbf{Dense Attention}, then its total complexity would be $24TND^2+4T^2N^2D$.

Transformer layer with \textbf{Space-Time Attention} contains 3 modules: space attention, time attention and FNN~\cite{SpaceTimeAttn}. Space attention requires $T(8ND^2+4N^2D)$ and time attention requires $N(8TD^2+4T^2D)$. Thus, the total complexity of a space-time layer is $32TND^2+4TND(N+T)$.

For \textbf{hierarchical modeling}, the input video is divided into $S$ segments of $T/S$ frame each~\cite{HierVL}, and processed by a video encoder with space-time attention. This gives effective complexity of each transformer layer to be $32TND^2+4TND(N+T/S)$.


\section{Qualitative Results}
\begin{figure*}[t]
\centering
\includegraphics[width=0.85\textwidth]{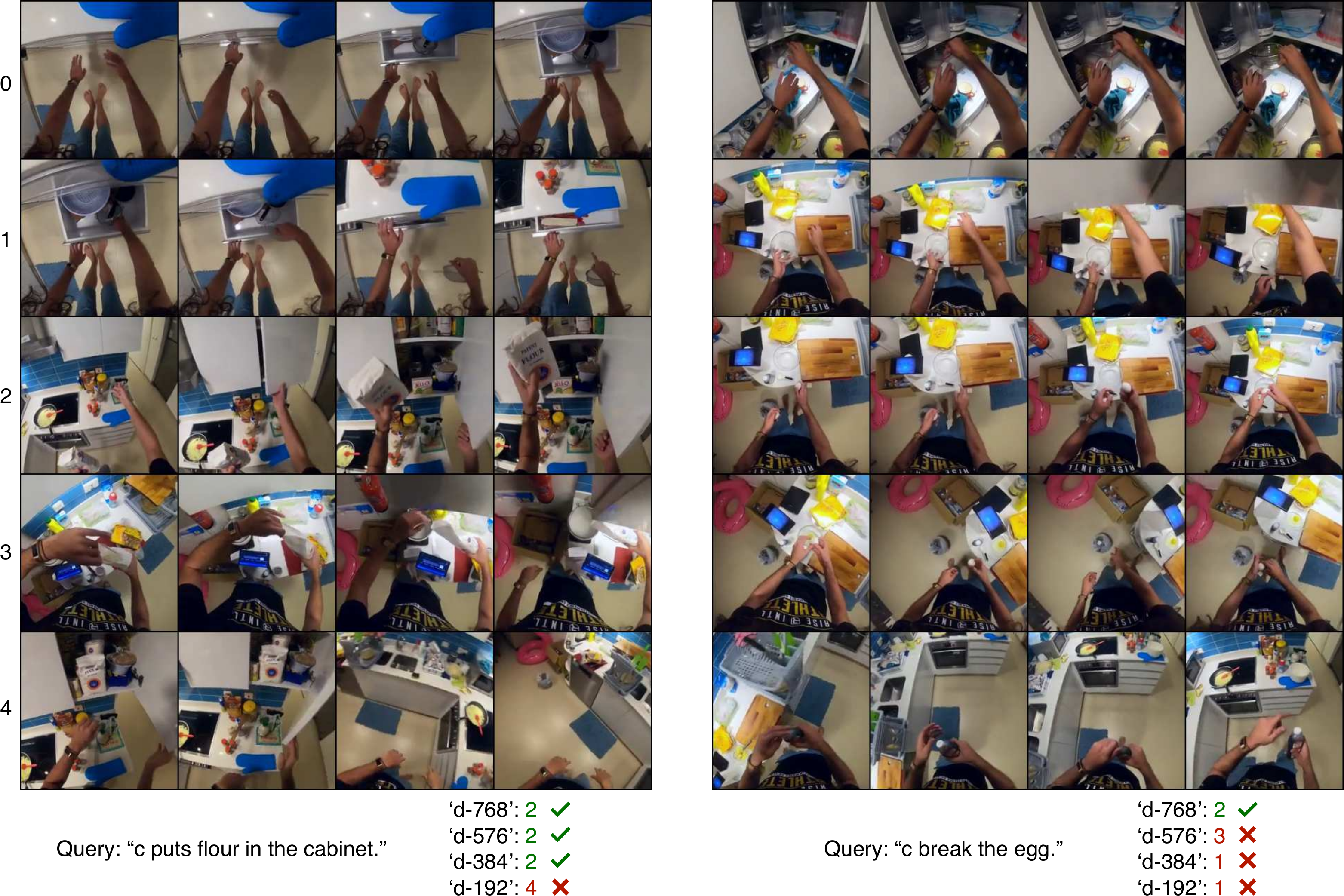} \\
\vspace{4mm}
\includegraphics[width=0.85\textwidth]{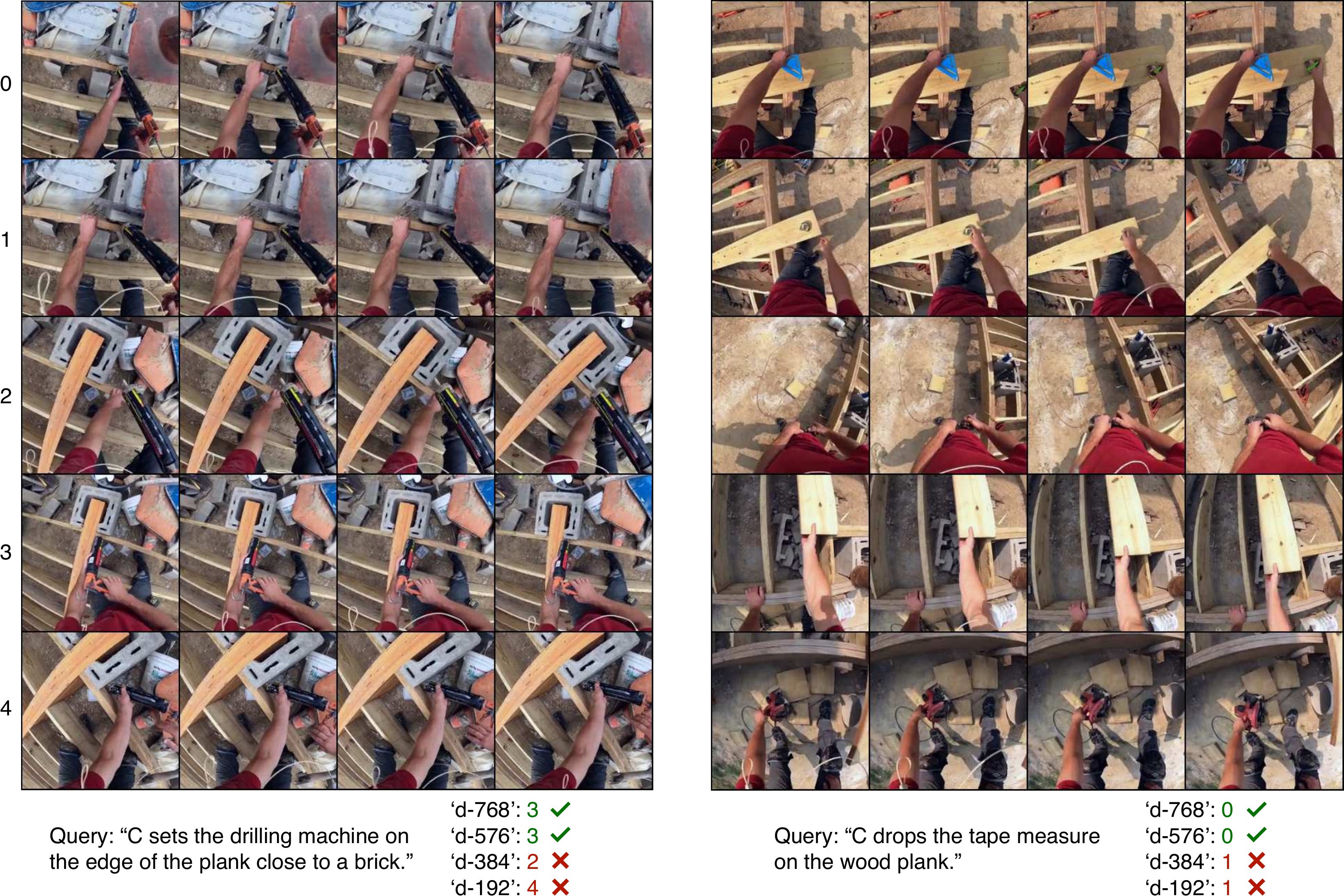}
\caption{We show four challenging examples from the EgoMCQ(intra) benchmark, each consisting of a text query and five candidate video clips. We also show the predictions made by our AdaVid-EgoVLP model under four different evaluation configurations. The results indicate that the model can do accurate fine-grained video analysis by adaptively increasing its compute.}
\label{fig:egomcq_vis}
\end{figure*}

\begin{figure*}[t]
\centering
\includegraphics[width=0.84\textwidth]{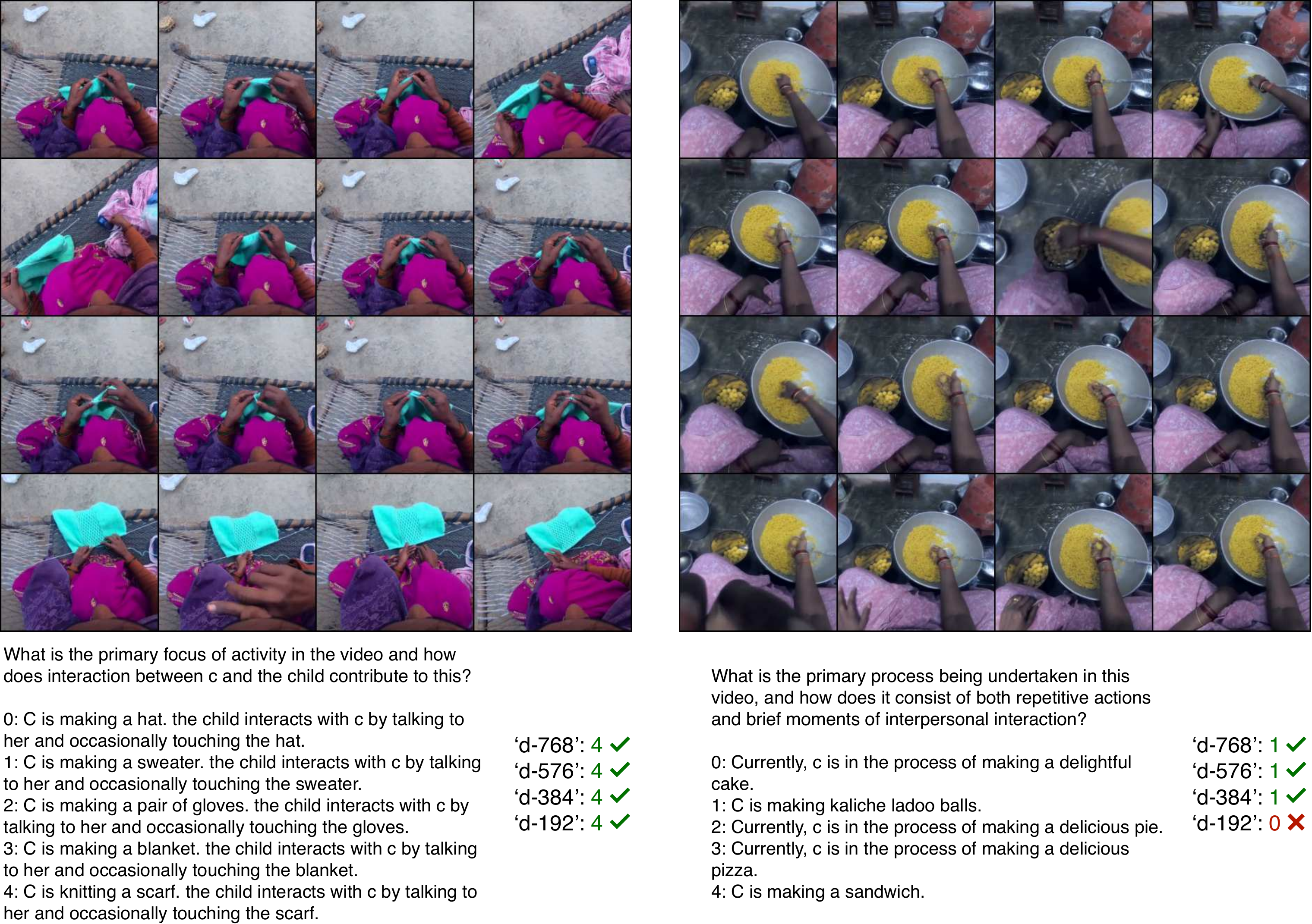} \\
\vspace{4mm}
\includegraphics[width=0.84\textwidth]{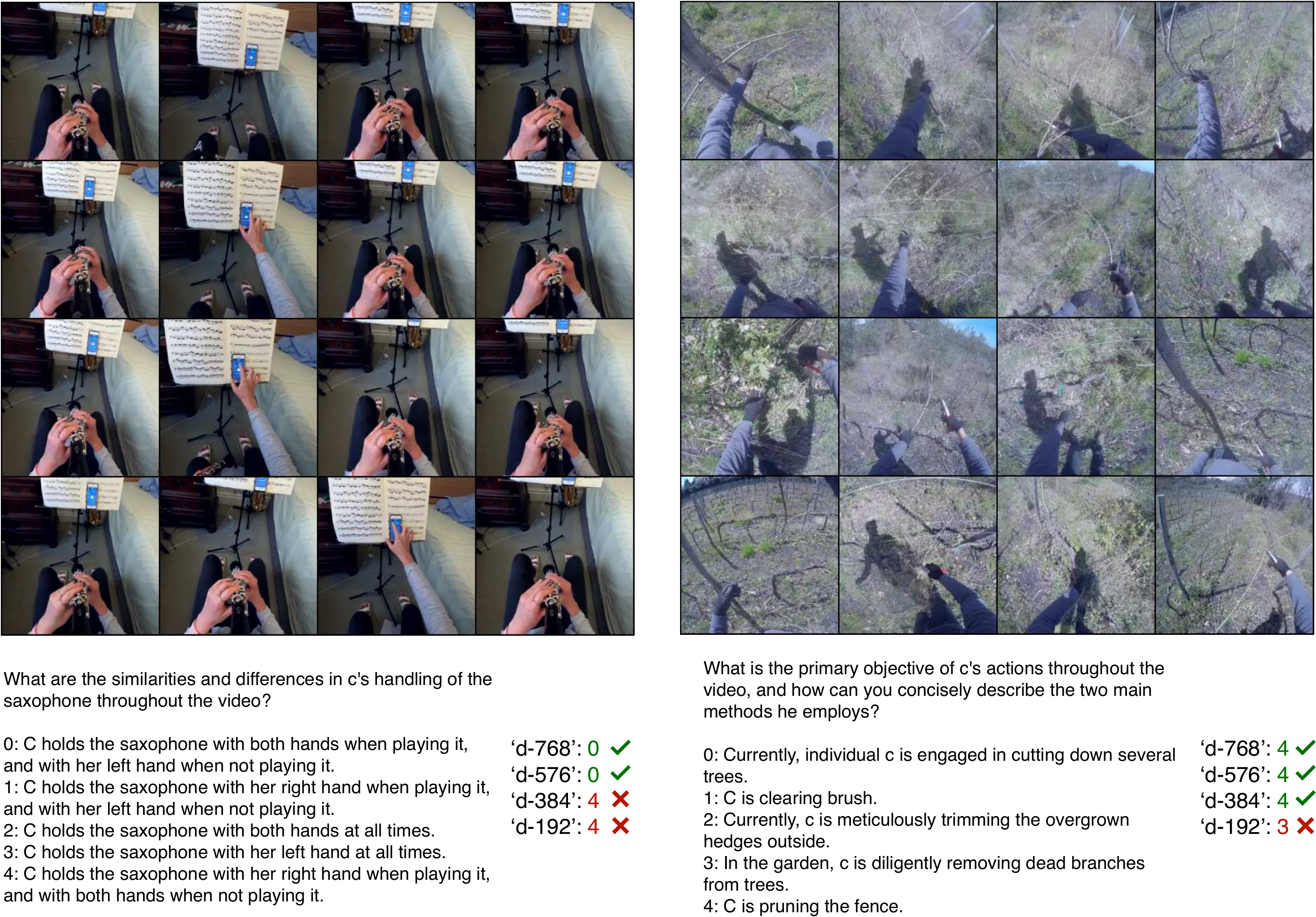}
\caption{We show four examples from the EgoSchema VideoQA benchmark, each consisting of a video and a question with 5 candidate answers. We also show the predictions made by our AdaVid-Agg model under four different evaluation configurations. The results indicate that the model can do long-form video analysis efficiently by adaptively increasing its compute.}
\label{fig:egoschema_vis}
\end{figure*}

In ~\Cref{fig:egomcq_vis} and ~\Cref{fig:egoschema_vis}, we show visualizations of our model's predictions under various evaluation configurations for EgoMCQ and EgoSchema benchmarks, respectively.

\section{AdaVid-Agg Architecture}
\begin{figure*}[t]
\centering
\includegraphics[width=0.8\linewidth]{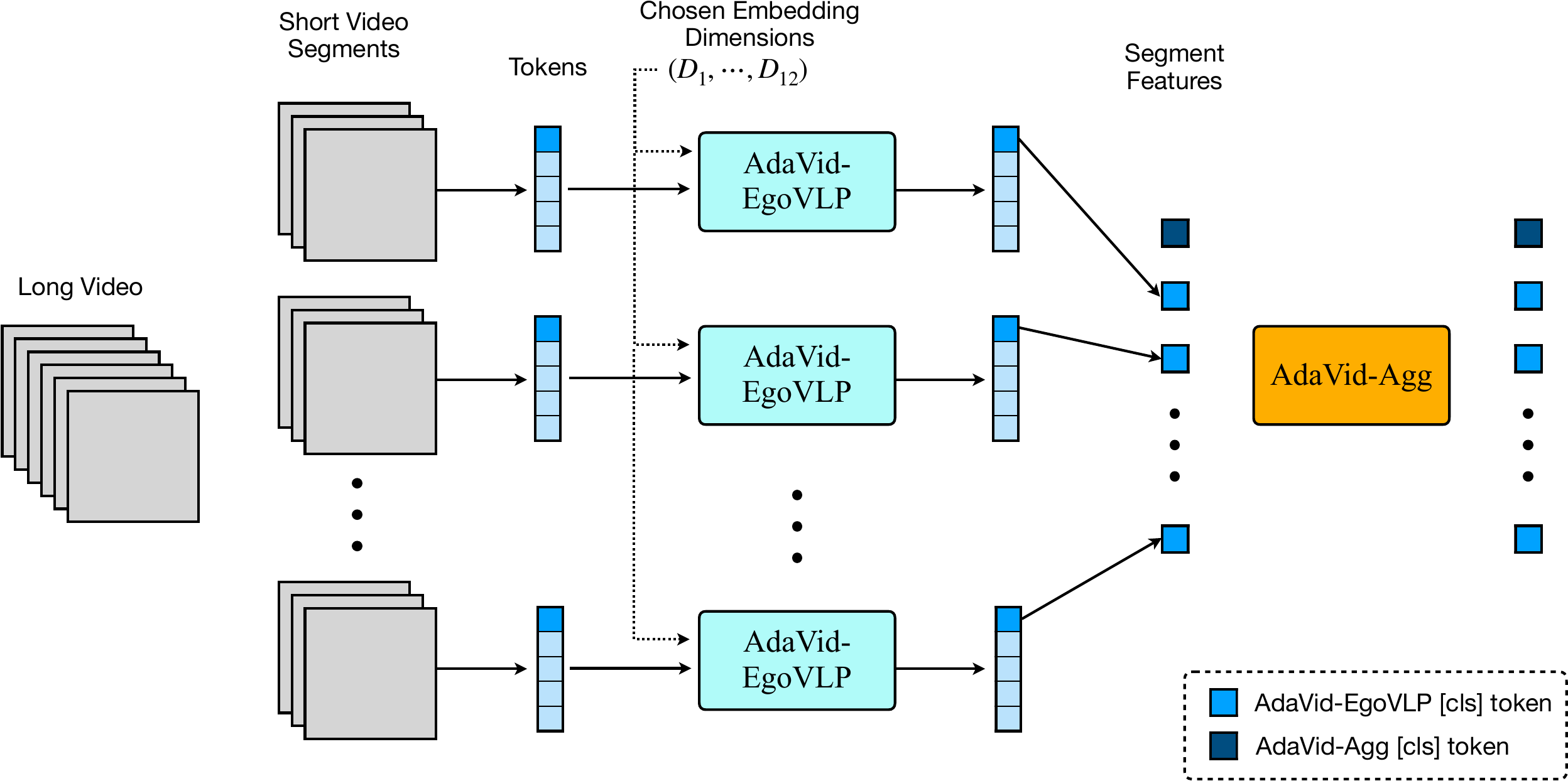}
\caption{\textbf{AdaVid-Agg}: A long video is divided into $S=16$ shorter segments and encoded using the pretrained AdaVid-EgoVLP model. The sequence of segment features is then processed by the lightweight AdaVid-Agg transformer model, which predicts a single feature representation for the entire video. It is important to note that AdaVid-EgoVLP is trained independently and remains frozen during the training of AdaVid-Agg.}
\label{fig:adavid_agg_arch}
\end{figure*}

~\Cref{fig:adavid_agg_arch} shows the architecture of the lightweight hierarchical AdaVid-Agg model. AdaVid-Agg is designed to encode a sequence of AdaVid-EgoVLP features into the feature representation for long videos. During the training of AdaVid-Agg, we extract features from AdaVid-EgoVLP using various evaluation configurations. This approach enables the entire pipeline to be compute adaptive during inference.


\end{document}